\documentclass[10pt,journal,compsoc]{IEEEtran}

\usepackage{times}
\usepackage{epsfig}
\usepackage{graphicx}
\usepackage{amsmath}
\usepackage{amssymb}
\usepackage[accsupp]{axessibility}

\usepackage{tabularx}
\usepackage{amssymb}
\usepackage{booktabs}
\usepackage{color}

\usepackage{multirow}
\usepackage[table,xcdraw]{xcolor}

\usepackage[pagebackref=true,breaklinks=true,colorlinks,bookmarks=false]{hyperref}
\usepackage{hyperref}

\ifCLASSOPTIONcompsoc

  \usepackage[nocompress]{cite}
\else
  \usepackage{cite}
\fi

\ifCLASSINFOpdf

\else

\fi

\begin{document}

\title{GyroFlow+: Gyroscope-Guided Unsupervised Deep Homography and Optical Flow Learning}








\author{Haipeng~Li,~\IEEEmembership{Student Member,~IEEE,}
        Kunming~Luo,
        Bing~Zeng, ~\IEEEmembership{Fellow,~IEEE}
        and~Shuaicheng~Liu,~\IEEEmembership{Member,~IEEE}
\IEEEcompsocitemizethanks{
\IEEEcompsocthanksitem Manuscript submitted at 06. Sept, 2022. This work was supported in part by the National Natural Science Foundation of China (NSFC) under Grants No. 61872067, No. 62031009 and No. 61720106004.

\IEEEcompsocthanksitem Haipeng Li, Bing Zeng and Shuaicheng Liu are with the School of Information and Communication Engineering, University of Electronic Science and Technology of China, Chengdu, Sichuan, 611731, China.

\IEEEcompsocthanksitem Kunming Luo is with Megvii Technology, China.

\IEEEcompsocthanksitem Corresponding author: Shuaicheng Liu (liushuaicheng@uestc.edu.cn)
}
}


\IEEEtitleabstractindextext{%
\begin{abstract}
Existing homography and optical flow methods are erroneous in challenging scenes, such as fog, rain, night, and snow because the basic assumptions such as brightness and gradient constancy are broken. To address this issue, we present an unsupervised learning approach that fuses gyroscope into homography and optical flow learning. Specifically, we first convert gyroscope readings into motion fields named gyro field. Second, we design a self-guided fusion module (SGF) to fuse the background motion extracted from the gyro field with the optical flow and guide the network to focus on motion details. Meanwhile, we propose a homography decoder module (HD) to combine gyro field and intermediate results of SGF to produce the homography. To the best of our knowledge, this is the first deep learning framework that fuses gyroscope data and image content for both deep homography and optical flow learning. To validate our method, we propose a new dataset that covers regular and challenging scenes. Experiments show that our method outperforms the state-of-the-art methods in both regular and challenging scenes. 
\end{abstract}

}

\maketitle

\IEEEdisplaynontitleabstractindextext

\IEEEpeerreviewmaketitle

\IEEEraisesectionheading{\section{Introduction}\label{sec:introduction}}

Homography estimation is a basic task in computer vision that has been widely used for a wide range of vision tasks, including image/video stitching~\cite{zaragoza2013projective}, digital video stabilization~\cite{liu2013bundled}, simultaneous localization and mapping (SLAM)~\cite{mur2015orb}, high dynamic range (HDR)~\cite{gelfand2010multi}, super resolution (SR)~\cite{ji2008robust,wang2020deep}, and noise reduction (NR)~\cite{cheng2021nbnet}. The homography matrix represents the perspective transformation between $2$ images of a certain plane. Traditional methods estimate the homography in multiple stages, including feature points detecting~\cite{lowe2004distinctive}, correspondence matching~\cite{cunningham2021k}, the direct linear transform~\cite{hartley2003multiple}, and outlier rejecting~\cite{fischler1981random}. However, these methods heavily rely on feature matches and fail in cases that lack high-quality features. Deep learning methods take a pair of images as input and directly produce a homography without matching key points, being more robust than the former ones. They can be divided into supervised and unsupervised methods that adopt synthetic data and photometric loss for training respectively. Some unsupervised algorithms~\cite{zhang2020content,ye2021motion,nguyen2018unsupervised} focus on learning homography under challenging cases, but they still somewhat partly rely on the image content.

Optical flow estimation is another fundamental yet essential computer vision task that has been widely applied in various applications such as object tracking~\cite{behl2017bounding}, visual odometry~\cite{campbell2004techniques}, and image alignments~\cite{kroeger2016fast}. The original formulation of the optical flow was proposed by Horn and Schunck~\cite{horn1981determining}, after which the accuracy of optical flow estimation algorithms has improved steadily. Early traditional methods minimize pre-defined energy functions with various priors and constraints~\cite{lucas1981iterative}. Deep learning methods directly learn the per-pixel regression through convolutional neural networks, which can be divided into supervised~\cite{dosovitskiy2015flownet, ranjan2017optical, teed2020raft} and unsupervised methods~\cite{ren2017unsupervised, luo2021upflow}. The former methods are primarily trained on synthetic data~\cite{dosovitskiy2015flownet, butler2012naturalistic} due to the lack of real-world ground-truth labels. In contrast, the later ones can be trained on abundant and diverse unlabeled data by minimizing the photometric loss between two images. Although existing methods achieve good results in multiple benchmarks~\cite{butler2012naturalistic,geiger2012we,menze2015object,dosovitskiy2015flownet}, they rely on image contents, requiring images to contain rich texture and similar illumination conditions. As a result, they often fail in challenging cases.

\begin{figure}[t]
\begin{center}
  \includegraphics[width=1\linewidth]{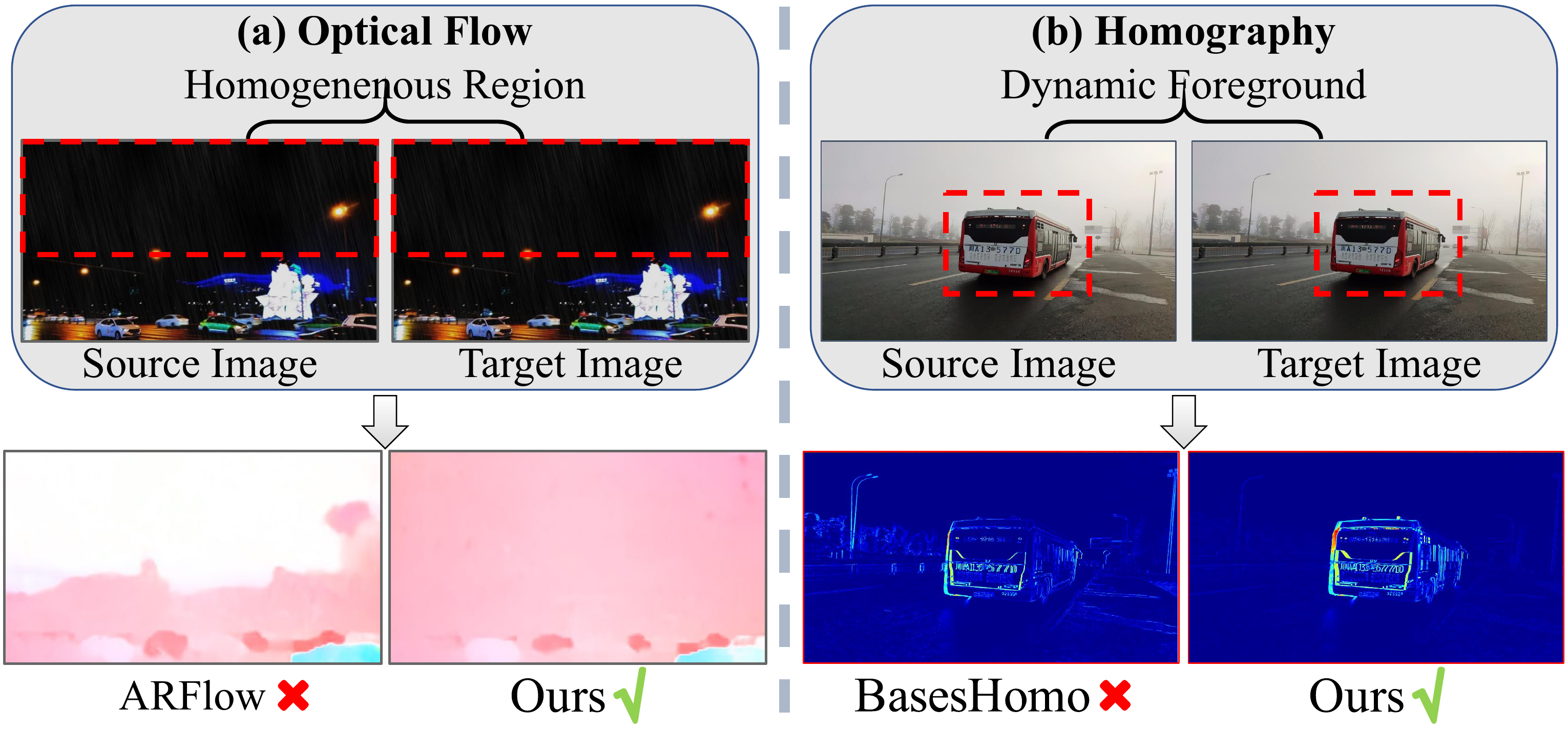}
\end{center}
  \caption{(a) Performance of optical flow methods in night scenes. The homogeneous region is highlighted in red boxes and the baseline method ARFlow~\cite{liu2020learning} fails to compute it while our method is capable of handling it. (b) Comparison of homography methods in foggy scenes. We illustrate the error heatmap between source and warped target image in the second row, the darker the image, the better the alignment. Due to the existence of challenging scenes and dynamic foregrounds, the recent method~\cite{ye2021motion} is unable to estimate a correct homography, while our method successfully tackles it.}
\label{fig:teaser}
\end{figure}

Different from the above methods, gyroscopes do not rely on image contents, which provide angular velocities in terms of roll, pitch, and yaw that can be converted into $3$D motion, widely used for system control~\cite{leland2006adaptive} and the human-computer interaction (HCI) of mobiles~\cite{gupta2016continuous}. Among all potential possibilities~\cite{bloesch2014fusion, li2018efficient, hwangbo2009inertial}, one is to fuse the gyro data for the motion estimation. Hwangbo~\emph{et al.} proposed to fuse gyroscope to improve the robustness of KLT feature tracking~\cite{hwangbo2009inertial}. Bloesch~\emph{et al.} fused gyroscope for the ego-motion estimation~\cite{bloesch2014fusion}. These attempts demonstrate that if the gyroscope is integrated correctly, the performance and the robustness of the method can be largely improved.   

Given camera intrinsic parameters, gyro readings can be converted into motion fields to describe background motion because it is confined to camera motion. It is engaging that gyroscopes do not require the image contents but still produce reliable background camera motion under conditions of poor texture or dynamic scenes. Therefore, gyroscopes can be used to improve the performance of homography and optical flow estimation in challenging scenes, such as poor texture or inconsistent illumination conditions.

In this paper, we propose GyroFlow+, a gyroscope-guided unsupervised deep homography and optical flow estimation method. We combine the advantages of image-based methods that recover motion details based on the image content with those of a gyroscope that provides reliable background camera motion independent of image contents. Specifically, we first convert gyroscope readings into gyro fields that describe background motion given the image coordinates and the camera intrinsic. Second, we estimate homography and optical flow with an unsupervised learning framework which includes a \textbf{S}elf-\textbf{G}uided \textbf{F}usion (SGF) module that supports the fusion of the gyro field during the image-based flow calculation, a \textbf{H}omography \textbf{D}ecoder (HD) that implements the gyro field with the feature maps and output of the SGF. Third, the gyro field is replaced by the produced homography to further improve the performance of optical flow. Thus, we realize the mutual promotion of optical flow and homography estimation. Fig.~\ref{fig:teaser} shows an example, where Fig.~\ref{fig:teaser} (a) represents the input of a night scene with poor image texture, the result on the left side is an image-based method, i.e., ARFlow~\cite{liu2020learning} that fails to compute the background motion in the sky which is the homogeneous region~\cite{truong2020glu}. The right side shows our optical flow result. As seen, both global motion and motion details can be retained. Fig.~\ref{fig:teaser} (b) is a foggy example, the image-based method, i.e., BasesHomo~\cite{ye2021motion} is not able to handle the challenging case where dynamic objects are included, while our homography result successfully aligns the inputs. 


To validate our method, we propose a dataset GHOF (\textbf{G}yroscope \textbf{H}omography \textbf{O}ptical \textbf{F}low) containing scenes under $5$ different categories with synchronized gyro readings, including one regular scene (RE) and four challenging cases as low light scenes (Dark), foggy scenes (Fog), rainy scenes (Rain) and snowy scenes (Snow). For quantitative evaluations, we further propose a benchmark, which includes accurate optical flow labels by the method~\cite{liu2008human} and homography ground-truth by the method~\cite{zhang2020content}, through extensive efforts. Note that existing flow datasets nor homography datasets, such as Sintel~\cite{butler2012naturalistic}, KITTI~\cite{geiger2012we,menze2015object}, and CAHomo~\cite{zhang2020content} cannot be used for the evaluation due to the absence of the gyroscope readings. To sum up, our main contributions are:

\begin{itemize}
    \item We propose the first deep learning framework that fuses gyroscope data into homography and optical flow learning.
    \item We propose a self-guided fusion module to realize the fusion of gyroscope and optical flow, and a homography decoder to implement gyro field into homography.
    \item We propose a dataset with accurate gyroscope data for the evaluation of homography and optical flow. Experiments show that our method outperforms existing methods.
\end{itemize}

\section{Related Work\label{sec:related}}

\subsection{Gyro-based Vision Applications}
Gyroscopes reflect the camera rotation. Many applications equipped with the gyroscope have been widely applied, including but not limited to video stabilization~\cite{karpenko2011digital}, image deblur~\cite{mustaniemi2019gyroscope}, optical image stabilizer (OIS)~\cite{la2015optical}, simultaneous localization and mapping (SLAM)~\cite{huang2018online}, ego-motion estimation~\cite{bloesch2014fusion}, gesture-based user authentication on mobile devices~\cite{guse2012gesture}, image alignment with OIS calibration~\cite{li2021deepois} and human gait recognition~\cite{zhang2004human}. The gyroscopes are important in mobile phones. The synchronization between the gyro readings and the video frames is important. Jia~\emph{et al.}~\cite{jia2013online} proposed gyroscope calibration to improve the synchronization. Bloesch~\emph{et al.}~\cite{bloesch2014fusion} fused optical flow and inertial measurements to deal with the drifting issue. In this work, we acquire gyroscope data from the bottom layer of the Android layout, i.e., Hardware Abstraction Layer (HAL), to achieve accurate synchronizations.

\subsection{Homography}
Our method is able to produce homography. Traditional methods estimate the homography in $3$ stages, including feature detecting~\cite{lowe2004distinctive,rublee2011orb}, correspondence matching~\cite{cunningham2021k}, and outlier rejecting~\cite{fischler1981random}. Recently, learning-based feature detecting and matching algorithms~\cite{yi2016lift,detone2018superpoint,tian2019sosnet} improve the performance and robustness to estimate homography. Furthermore, outlier rejecting methods~\cite{fischler1981random,barath2019magsac,barath2020magsac++} promote the effectiveness and stability of the algorithm in complex scenes, such as multi-plane, parallax, dynamic foreground, etc. In recent years, deep learning methods have emerged to improve the performance and efficiency of producing homography which can be divided into supervised~\cite{detone2016deep,le2020deep,shao2021localtrans,cao2022iterative} and unsupervised methods~\cite{nguyen2018unsupervised,zhang2020content,ye2021motion,kharismawati2020cornet,hong2022unsupervised}.

The supervised methods randomly sample homography, warping an image to form a pair of training data with the ground-truth label, however, their method fails in real-world scenes due to the lack of generalization capability. To address the issue, unsupervised methods are proposed by collecting real-world images and minimizing photometric loss. To be more robust, ~\cite{zhang2020content} proposes an explicit outlier mask to remove undesired regions, ~\cite{ye2021motion} proposes to limit the rank of feature maps, to constrain a feature-identity loss and generate an $8$ weighted basis to implicitly improve the results. In order to guide the network to focus on the dominant plane and improve the receptive field, ~\cite{hong2022unsupervised} proposes a GAN loss to identify the primary plane and a Transformer encoder to realize the coarse-to-fine manner. In this work, we estimate a homography based on the gyroscope data to solve the above problems.

\subsection{Optical Flow}
Our method is related to optical flow estimation. Traditional methods minimize the energy function between image pairs to compute an optical flow~\cite{lucas1981iterative}. Recent deep approaches can be divided into supervised~\cite{dosovitskiy2015flownet, ranjan2017optical, teed2020raft} and unsupervised methods~\cite{ren2017unsupervised, luo2021upflow}. 

\begin{figure*}[t]
\begin{center}
  \includegraphics[width=0.95\linewidth]{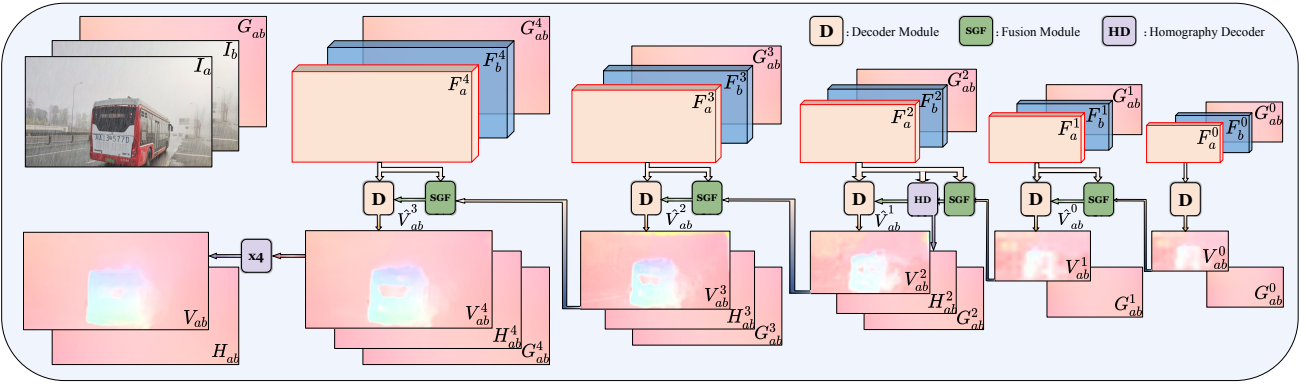}
\end{center}
  \caption{The overview of our algorithm. It consists of a pyramid encoder and a pyramid decoder. For each pair of frames $I_{a}$ to $I_{b}$, our encoder extracts features at different scales. The decoder includes two or three modules, at each layer $l$, $\mathbf{SGF}$ functions to fuse a gyro field $G_{ab}^{l}$ and an optical flow $V_{ab}^{l}$ to produce a fused flow $\hat{V_{ab}^{l}}$ as input to $\mathbf{D}$, which estimates an optical flow to the next layer. At a certain layer, $\mathbf{HD}$ inputs features, gyro field, and weight map from $\mathbf{SGF}$ to produce the homography $H_{ab}^{l}$.}
\label{fig:pipeline}
\end{figure*}

Supervised methods require labeled ground-truth to train the network. FlowNet~\cite{dosovitskiy2015flownet} first proposed to train a fully convolutional network on the synthetic dataset FlyingChairs. To deal with the large displacement scenes, SpyNet~\cite{ranjan2017optical} introduced a coarse-to-fine pyramid network. PWC-Net~\cite{sun2018pwc}, LiteFlowNet~\cite{hui2018liteflownet}, IRR-PWC~\cite{hur2019iterative} designed lightweight and efficient networks by warping features, computing cost volumes, and introducing residual learning for iterative refinement with shared weights. Recently, RAFT~\cite{teed2020raft} achieved state-of-the-art performance by constructing a pixel-level correlation volume and using a recurrent network to estimate optical flow. 

Unsupervised methods do not require ground-truth annotations. DSTFlow~\cite{ren2017unsupervised} and Back2Basic~\cite{jason2016back} are pioneers for unsupervised optical flow estimation. Several works~\cite{meister2018unflow, liu2019ddflow, wang2018occlusion, liu2020learning} focus on dealing with the occlusion problem by forward-backward occlusion checking, range-map occlusion checking, data distillation, and augmentation regularization loss. Other methods concentrate on optical flow learning by improving image alignment, including the census loss~\cite{meister2018unflow}, formulation of multi-frames~\cite{janai2018unsupervised}, epipolar constraints~\cite{zhong2019unsupervised}, depth constraints~\cite{yin2018geonet}, feature similarity constraints~\cite{im2020unsupervised}, and occlusion inpainting~\cite{liu2021oiflow}. UFlow~\cite{jonschkowski2020matters} proposed a unified framework to systematically analyze and integrate different unsupervised components. Recently, UPFlow~\cite{luo2021upflow} proposed a neural upsampling module and pyramid distillation loss to improve the upsampling and learning of the pyramid network, achieving state-of-art performance. 

However above methods may not work well under challenging scenes, such as dark, rain, and fog environments. Zheng~\emph{et al.} proposed a data-driven method that establishes a noise model to learn optical flow from low-light images~\cite{zheng2020optical}. Li~\emph{et al.} proposed a RainFlow, which includes $2$ modules to handle the rain veiling effect and rain streak effect respectively, to produce optical flow in the heavy rain~\cite{li2019rainflow}. Yan~\emph{et al.} proposed a semi-supervised network that converts foggy images into clean images to deal with dense foggy scenes~\cite{yan2020optical}. In this paper, we build our GyroFlow upon unsupervised components with the fusion of gyroscope to cover both regular and challenging scenes.

\subsection{Gyro-based Motion Estimation}
Hwangbo~\emph{et al.} proposed an inertial-aided KLT feature tracking method to handle the camera rolling and illumination change~\cite{hwangbo2009inertial}. Bloesch~\emph{et al.} presented a method for fusing optical flow and inertial measurements for robust ego-motion estimation~\cite{bloesch2014fusion}. Li~\emph{et al.} proposed a gyro-aided optical flow estimation method to improve the performance under fast rotations~\cite{li2018efficient}. Specifically, they produce a sparse optical flow that ignores foreground motion. However, none of them took challenging scenes into account nor used neural networks to fuse gyroscope data for optical flow improvement. 

In this work, including producing homography, dense optical flow and taking rolling-shutter effects into account, we propose a deep learning solution to improve homography and optical flow estimations.

\section{Algorithm}
It is complementary to estimate optical flow and homography for giving the gyro field. On the one hand, a mask for filtering outliers can be produced in the process of estimating the optical flow. On the other hand,  after using homography field to replace gyro field, the results of optical flow can be enhanced. To achieve this, our method is built upon convolutional neural networks that input a gyro field $G_{ab}$ and two frames $I_{a}$, $I_{b}$ to estimate the homography $H_{ab}$ and optical flow $V_{ab}$ that respectively describes the camera motion and the motion for every pixel from $I_{a}$ towards $I_{b}$ as:
\begin{equation}
H_{ab}, V_{ab} = \mathcal{F}_{\theta}\left(G_{ab}, I_a, I_b\right),
\end{equation}
where $\mathcal{F}$ is our network with parameter $\theta$.

Fig.~\ref{fig:pipeline} illustrates our pipeline. Firstly, the gyro field $G_{ab}$ is produced by the gyroscope readings between the relative frames $I_{a}$ and $I_{b}$ (Sec.~\ref{sec:gyro_field}), then it is concatenated with the two frames to be fed into the network to produce a homography $H_{ab}$ and an optical flow $V_{ab}$ between $I_{a}$ and $I_{b}$. Our network consists of two stages. For the first stage, we extract feature pairs at different scales. For the second stage, we use the decoder $\mathbf{D}$, the self-guided fusion module $\mathbf{SGF}$ (Sec.~\ref{sec:SFG}), and homography decoder $\mathbf{HD}$ to produce homography, optical flow in a coarse-to-fine manner.

Our decoder $\mathbf{D}$ is the same as UPFlow~\cite{luo2021upflow} which consists of the feature warping~\cite{sun2018pwc}, the cost volume construction~\cite{sun2018pwc}, the cost volume normalization~\cite{jonschkowski2020matters}, the self-guided upsampling~\cite{luo2021upflow}, and the parameter sharing~\cite{hur2019iterative}. In summary, the second pyramid decoding stage can be formulated as:

\begin{equation}
\begin{aligned}
\hat{V}_{ab}^{i-1},M_{ab}^{i-1} &=\mathbf{SGF}\left(F_{a}^{i}, F_{b}^{i},V_{ab}^{i-1},G_{ab}^{i-1}\right), \\
{H}_{ab}^{i} &=\mathbf{HD}\left(F_{a}^{i}, F_{b}^{i},M_{ab}^{i-1},G_{ab}^{i-1}\right), \\
V_{ab}^{i} &=\mathbf{D}\left(F_{a}^{i}, F_{b}^{i}, \hat{V}_{ab}^{i-1}\right),
\end{aligned}
\end{equation}
where $i$ represents the number of pyramid levels, $F_{a}^{i}$, $F_{b}^{i}$ are extracted features from $I_{a}$ and $I_{b}$ at the $i$-th pyramid level. In the $i$-th layer, $\mathbf{SGF}$ takes image features $F_{a}^{i}$, $F_{b}^{i}$ from the feature pyramid, the output $V_{ab}^{i-1}$ of decoder $\mathbf{D}$ from the last layer and the downscale gyro field $G_{ab}^{i-1}$ as inputs, then it produces a fusion result $\hat{V}_{ab}^{i-1}$ and a weight map $M_{ab}^{i-1}$. The $\mathbf{HD}$ inputs features $F_{a}^{i}$, $F_{b}^{i}$, the weight map $M_{ab}^{i-1}$ and gyro field $G_{ab}^{i-1}$ and produces the homography ${H}_{ab}^{i}$. The $\mathbf{D}$ takes image features $F_{a}^{i}$, $F_{b}^{i}$ and the fusion result $\hat{V}_{ab}^{i-1}$ as inputs and outputs a flow $V_{ab}^{i}$. Specifically, the output flow is directly upsampled at the last layer. Next, we first describe how to convert the gyro readings into a gyro field in Sec.~\ref{sec:gyro_field}, then introduce our $\mathbf{SGF}$ module in Sec.~\ref{sec:SFG}, and finally present our $\mathbf{HD}$ in Sec.~\ref{sec:homo_module}.

\subsection{Gyro Field}\label{sec:gyro_field}

\begin{figure}[t]
\begin{center}
  \includegraphics[width=1\linewidth]{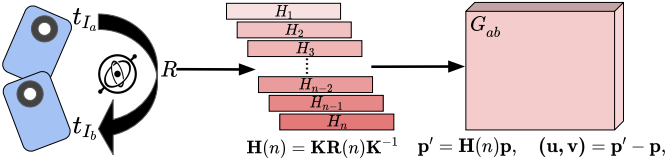}
\end{center}
  \caption{The pipeline of generating gyro field. Given timestamps $t_{I_a}$ and $t_{I_{b}}$, gyroscope readings can be read out to compute an array of rotation matrices $R=\left(R_{1} \ldots R_{n}\right)^\mathsf{T}$. We then convert the rotation array into the homography array that projects pixels $p$ of the first image into $p^{\prime}$, yielding a gyro field $G_{ab}$.}
\label{fig:gyro_pipe}
\end{figure}

We obtain gyroscope readings from mobile phones that are widely available and easy to access. For mobile phones, gyroscopes reflect camera rotations. We compute rotations by compounding gyroscope readings that include 3-axis angular velocities and timestamps. In particular, compared to previous work~\cite{karpenko2011digital, kundra2014bias, mustaniemi2019gyroscope} that read gyro readings from the API, we directly read them from HAL of Android architecture to avoid the non-trivial synchronization problem that is critical for gyro accuracy. Between frames $I_a$ and $I_b$, the rotation vector $n=\left(\omega_{x}, \omega_{y}, \omega_{z}\right) \in \mathbb{R}^{3}$ is computed according to method~\cite{karpenko2011digital}, then the rotation matrix $R(t)\in SO(3)$ can be produced by Rodrigues Formula~\cite{dai2015euler}.

In the case of a global shutter camera, e.g., the pinhole camera, a rotation-only homography can be computed as:
\begin{equation}
\mathbf{H(t)}=\mathbf{K} \mathbf{R}(t) \mathbf{K}^{-1}\label{eq:globalH},
\end{equation}
 where $K$ is the camera intrinsic matrix, $t$ represents the time from the first frame $I_{a}$ to the second frame $I_{b}$, and $\mathbf{R}(t)$ denotes the camera rotation from $I_a$ to $I_b$.

For a rolling shutter camera that most mobile phones adopt, each scanline of the image is exposed at a slightly different time, as illustrated in Fig~\ref{fig:gyro_pipe}. Therefore, Eq.~(\ref{eq:globalH}) is not applicable anymore, since every row of the image should have a different orientation. In practice, it is not necessary to assign each row with a rotation matrix. We group several consecutive rows into a row patch and assign each patch with a rotation matrix. The number of row patches depends on the number of gyroscope readings per frame.


Here, the homography between the $n$-th row at frame $I_a$ and $I_b$ can be modeled as:

\begin{equation}
\mathbf{H}_{n}(t)=\mathbf{K} \mathbf{R}\left(t_{b}\right) \mathbf{R}^{\top}\left(t_{a}\right) \mathbf{K}^{-1},
\end{equation}
where the $n$ is the index of row patches, $\mathbf{H}_{n}(t)$ denotes the homography of the $n$-th row patch from $I_{a}$ to $I_{b}$, and $\small\mathbf{R}\left(t_{b}\right) \mathbf{R}^{\top}\left(t_{a}\right)$ can be computed by accumulating rotation matrices from $t_a$ to $t_b$.

In our implementation, we regroup the image into $14$ patches that compute a homography array containing $14$ horizontal homography between two consecutive frames. Furthermore, to avoid the discontinuities across row patches, we convert the array of homography into an array of $4$D quaternions~\cite{zhang1997quaternions} and then apply the spherical linear interpolation (SLERP) to interpolate the camera orientation smoothly, yielding a smooth homography array. As shown in Fig~\ref{fig:gyro_pipe}, we use the homography array to transform every pixel $p$ to $p^{\prime}$, and subtract $p^{\prime}$ from $p$ as:

\begin{equation}
\mathbf{p}^{\prime}=\mathbf{H(n)} \mathbf{p}, \quad \mathbf{(u,v)}=\mathbf{p}^{\prime}- \mathbf{p},
\label{eq:homo2flow}
\end{equation}
computing offsets for every pixel produces a gyro field $G_{ab}$.

\subsection{Self-guided Fusion Module}\label{sec:SFG}
\begin{figure}[t]
\begin{center}
  \includegraphics[width=1\linewidth]{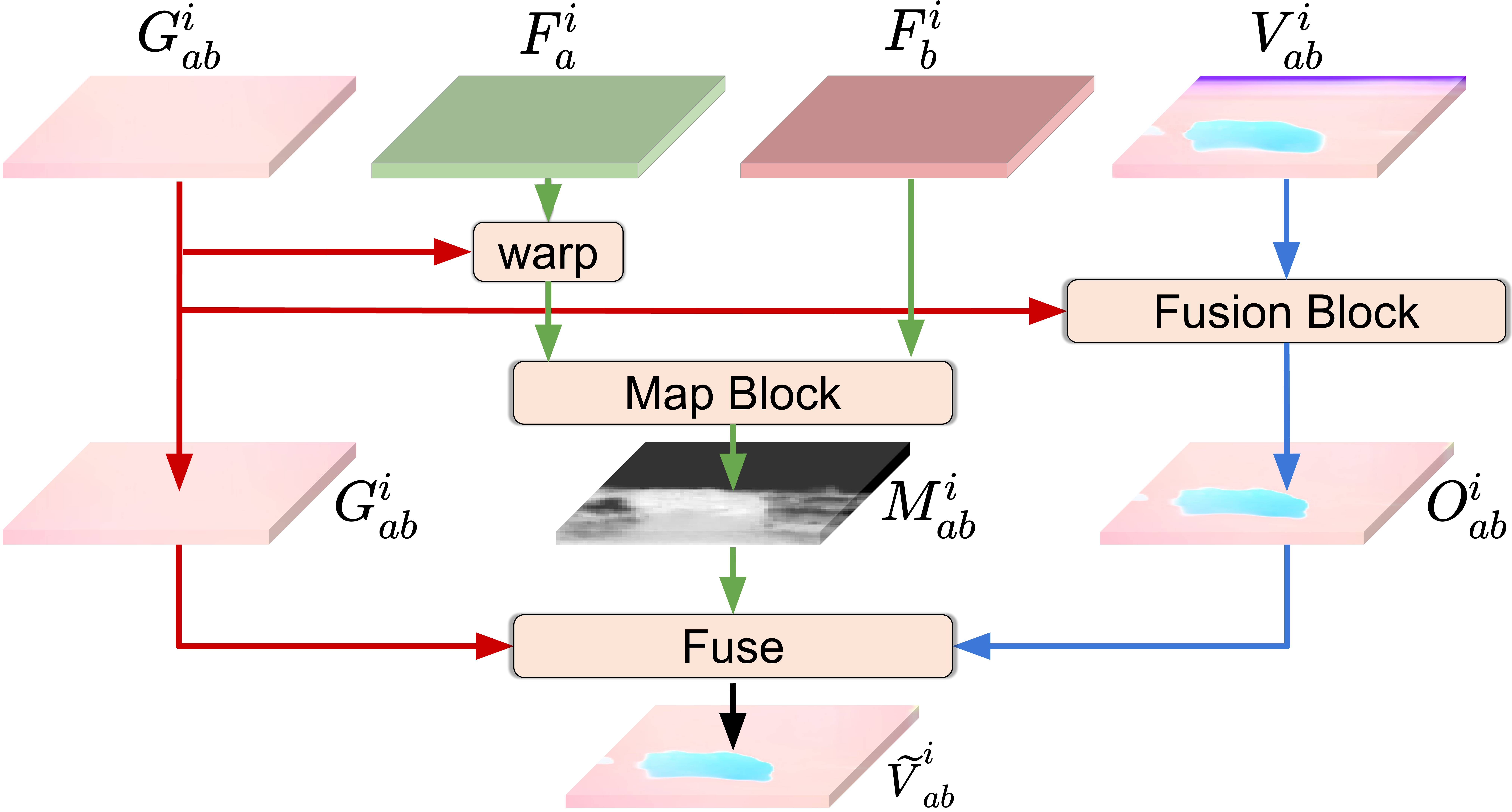}
\end{center}
  \caption{Illustration of our self-guided fusion module (SGF). For a specific layer $i$, we use $2$ blocks to independently produce the fusion map $M^{i}_{ab}$ and the fusion flow $O^{i}_{ab}$, then we generate the output $\widetilde{V}_{ab}^{i}$ by Eq.~\ref{fuse_eq}.}
\label{fig:SGF}
\end{figure}

Fig.~\ref{fig:teaser} (a) denotes the input images. The left side is the output of the ARFlow~\cite{liu2020learning}, an unsupervised optical flow approach, where only the motion of moving objects is roughly produced. As image-based optical flow methods count on image contents for the registration, they are prone to be erroneous in challenging scenes, such as textureless scenarios, dense foggy environments~\cite{yan2020optical}, dark~\cite{zheng2020optical} and rainy scenes~\cite{li2019rainflow}. To combine the advantages of the gyro field and the image-based optical flow, we propose a self-guided fusion module (SGF). On the right side, with the gyro field, our result is much better compared with the ARFlow~\cite{liu2020learning}. The architecture of our SGF is shown in Fig.~\ref{fig:SGF}.

 \textbf{Explicit Fusion.} Given the input features of image $I_a$ and $I_b$ at the $i$-th layer as $F^{i}_{a}$ and $F^{i}_{b}$. $F^{i}_{a}$ is warped by the gyro field $G^{i}_{ab}$, which is the forward flow from feature $F^{i}_{a}$ to $F^{i}_{b}$. Then the warped feature is concatenated with $F^{i}_{b}$ as inputs to the map block, yielding a fusion map $M^{i}_{ab}$ that ranges from $0$ to $1$. Note that, in $M^{i}_{ab}$, those background regions which can be aligned with the gyro field are prone to be zeros, while the rest areas are distributed with different weights. We realize an explicit fusion procedure by using the fusion map $M^{i}_{ab}$. 
 
 \begin{figure}[t]
\begin{center}
  \includegraphics[width=1\linewidth]{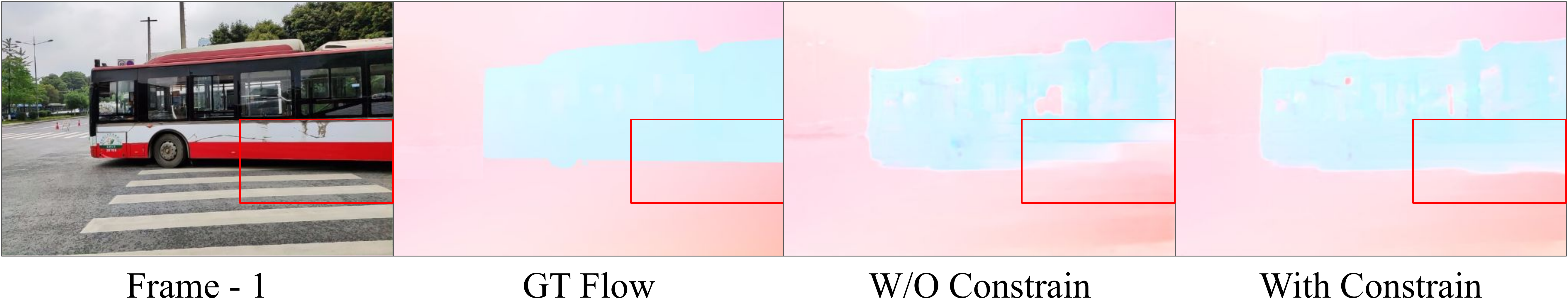}
\end{center}
  \caption{The first column is an input frame under regular scenarios, the homogeneous region is marked in red boxes. The second column shows the ground-truth optical flow. The output of $\mathbf{SGF}$ without constraint is illustrated in the third column, foreground regions overlaid by background motion. The last column shows the constrained result.}
\label{fig:Homogeneous_Region}
\end{figure}

\textbf{Constrained Explicit Fusion.} Notably, the distribution of the fusion map $M^{i}_{ab}$ is not constrained, which means that the network is completely free to choose the weights for every pixel. As a result, it causes a phenomenon where the network is over-dependent on the gyro field. Fig.~\ref{fig:Homogeneous_Region} shows an example, for the homogeneous regions in the foreground objects, such as solid color planes on cars, the network is more likely to select the gyro field. For these regions, instead of the gyro field, the foreground optical flow should be chosen. To address the problem, we further propose a global-to-local strategy, as the gyro field represents the global motion of the camera, it is natural to fuse more gyroscope information in the global layers, i.e., the top pyramid layer. While the bottom pyramid layer represents the local motion, so there ought to be less gyroscopic information. More specifically, for each pyramid layer, we separative restrict the ability of the network to fetch the gyro field, the more bottom the pyramid layer is, the more we limit its ability to acquire gyroscopes. We achieve this by manually setting the intervals of fusion map $M^{i}_{ab}$ as,
\begin{equation}
\beta_{5} < M^{4} \leq \beta_{4} \leq M^{3} \leq \beta_{3} \leq M^{2} \leq \beta_{2} \leq M^{1} \leq \beta_{1},
\label{map_W}
\end{equation}
where $\beta_{i}$ represents the upper weight of the gyro field the network can get at the $i$-th layer. Intuitively, each component is thus responsible for a different range of gyroscope content. In particular, component $i = 1$ accounts for the maximum degrees of freedom to fetch the gyro field, while component $i = 4$ models the biggest constraints. In order to enforce the constraint~\ref{map_W}, we predict a feature $h_{i} \in \mathbb{R}$ that maps to the given range as, 
\begin{equation}
M^{i}=\beta_{i}+\left(\beta_{i+1}-\beta_{i}\right) \operatorname{Sigmoid}\left(h_{i}\right).
\end{equation}


\textbf{Implicit Fusion.} Next, we input the gyro field $G^{i}_{ab}$ and optical flow $V^{i}_{ab}$ to the fusion block that computes a fusion flow $O^{i}_{ab}$. The fusion block consists of a series of convolutions blocks, $G^{i}_{ab}$ and $V^{i}_{ab}$ are stacked in the channel dimension to be fed into it, producing a new $2$ channels flow tensor $O^{i}_{ab}$. Finally, we fuse the $G^{i}_{ab}$ and $O^{i}_{ab}$ with $M^{i}_{ab}$ to guide the network to focus on the moving foreground regions. The process can be described as:

\begin{equation}
    \widetilde{V}_{ab}^{i}=M_{ab}^{i} \odot O_{ab}^{i}+\left(1-M_{ab}^{i}\right) \odot G_{ab}^{i},
    \label{fuse_eq}
\end{equation}
where $\widetilde{V}_{ab}^{i}$ is the output of our SGF module and $\odot$ denotes the element-wise multiplier. As illustrated in Fig.~\ref{fig:Homogeneous_Region}, the flow in homogenous regions is improved with our constrained strategy.

\subsection{Homography Decoder}\label{sec:homo_module}

 \begin{figure}[h]
\begin{center}
  \includegraphics[width=1\linewidth]{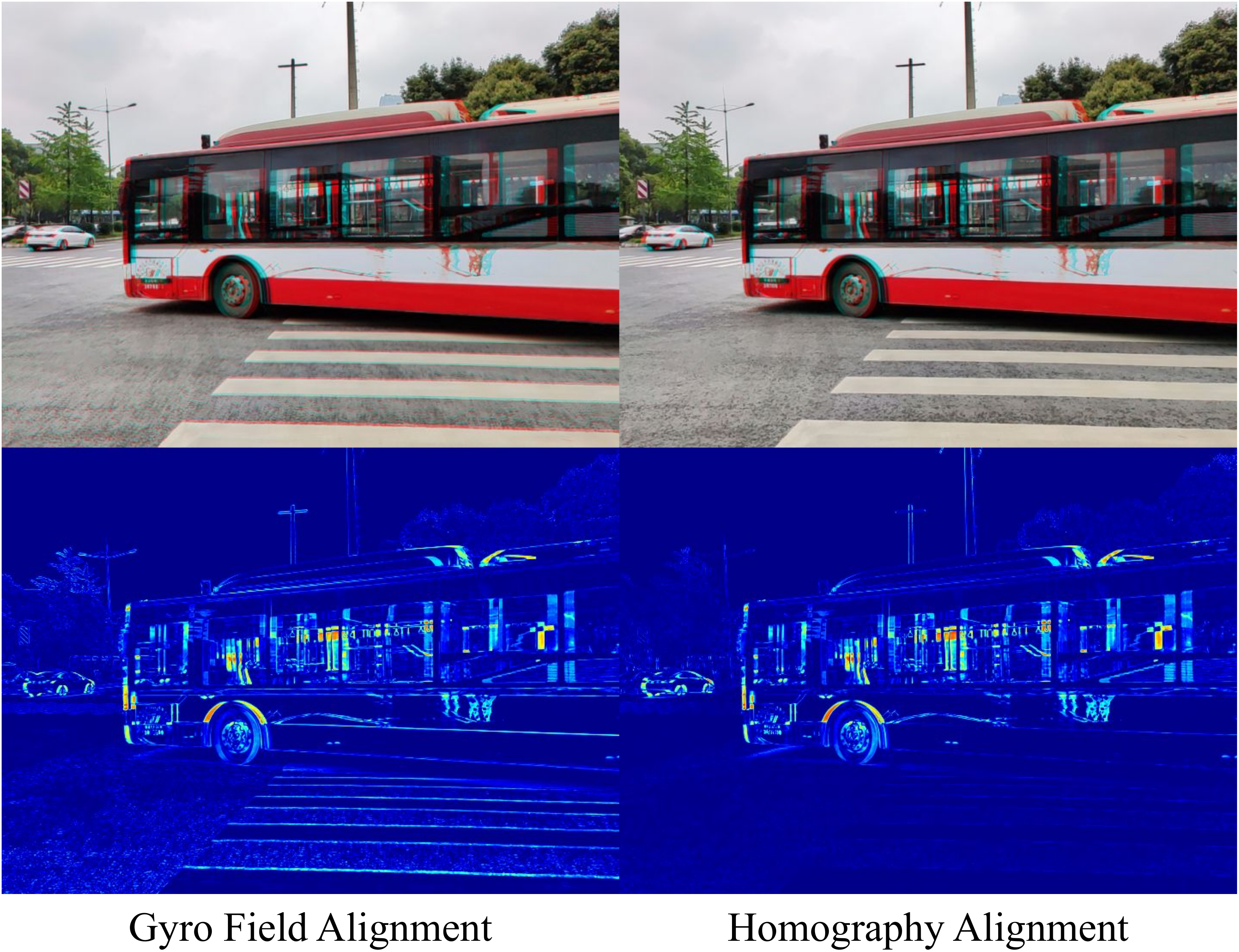}
\end{center}
  \caption{Visualization of alignment results. In the first row, we superimpose the warped source image and the target image, where misaligned pixels are visualized as colored ghosts. In order to highlight the details, the second row shows error maps, the darker the image, the better the alignment. The left side illustrates the output of the gyro field where the background region is well aligned because it is far from the camera center and is less affected by translation, while foreground regions are under-aligned. The right side is the output of $\mathbf{HD}$ where both foreground and background regions are aligned.}
\label{fig:homo_decoder}
\end{figure}

 \begin{figure}[h]
\begin{center}
  \includegraphics[width=1\linewidth]{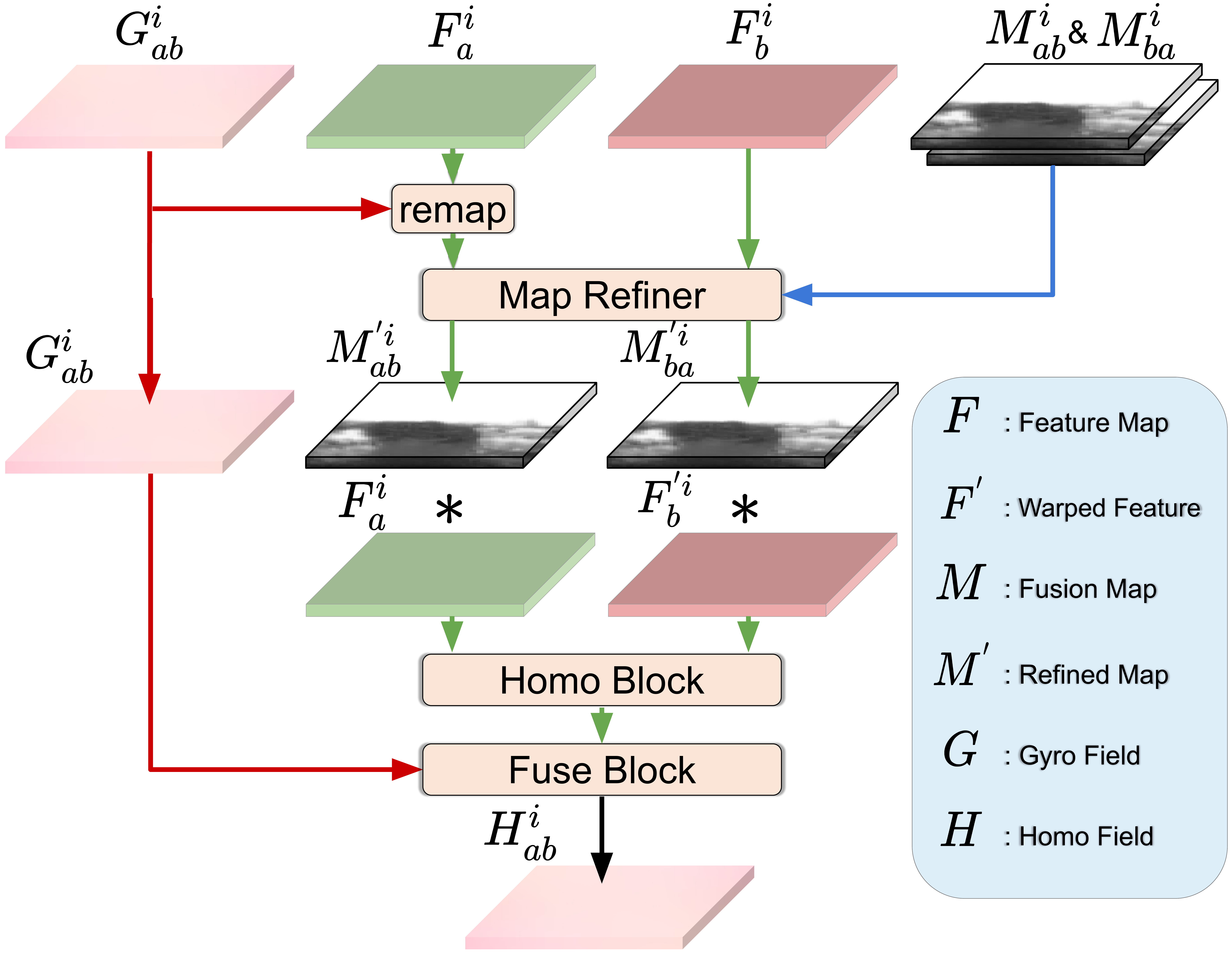}
\end{center}
  \caption{Illustration of our homography decoder(HD). For a specific layer $i$, we use $3$ blocks to independently refine the fusion map $M^{i}_{ab}$ and implement the gyro field $G^{i}_{ab}$, then we generate the output ${H}_{ab}^{i}$ as described in Sec.~\ref{sec:homo_module}.}
\label{fig:homo_decoder_vis}
\end{figure}

In our framework, the gyro field~\ref{sec:gyro_field} is essential, because it not only represents a reliable global motion but also it helps to produce a discriminative weight map to identify the foreground and background regions as illustrated in Fig.~\ref{fig:SGF-quatitative}. However, there still exist some minor defects, the most severe and challenging of which is the lack of translation information. In certain scenes, such as parallax, translation transformation, and multi-planar scenes, the alignment of the gyro field would be inferior to that of the complete homography. In Fig.~\ref{fig:homo_decoder}, we show an example that is common in nature, holding a camera across an intersection, which contains both rotational motion, translational motion, and parallax transformation. Two adjacent frames are selected, the first column shows the alignment result of gyro field and the second column illustrates the alignment result of homography. In order to highlight the alignment performance, two visualizations are leveraged, in the first row we exchange the red channels of the two frames so that the unaligned regions will be shown as red ghosts, and in the second row, we minus the two frames and convert them into an alignment heat map, where the bluer represents better alignment results and the redder represents worse ones. We find that the two methods achieve comparable alignment results in the background area, as the background is far from the camera center which is less affected by translation, however, if we zoom in and carefully check, the homography alignment result is still better. On the other hand, for regions close to the camera, the performance of homography is significantly better.

\textbf{Refine Fusion Map.} Predicting a homography is non-trivial as mentioned in ~\cite{hong2022unsupervised,ye2021motion,zhang2020content}. One of the most important components is the RANSAC-like mask which rejects outliers in the feature map and correspondences, furthermore, some methods~\cite{zhang2020content,hong2022unsupervised} can highlight important features or correspondences. Fortunately, our Fusion Map $M_{ab}^{i}$ produced by the SGF can somehow be a suitable starting point. In order to reduce the difficulty of learning and accelerate the convergence, the Fusion Map $M_{ab}^{i}$ is fed into the Homography Decoder. From our experiments, we find that $M_{ab}^{i}$ can roughly identify the foreground and background regions, and the weights in the foreground regions are remarkably smaller than the background regions as illustrated in Fig.~\ref{fig:SGF-quatitative}, however, the mean of the weight in foreground regions could sometimes be more than $0.2$ that is not satisfying the expectations. To address the issue, we propose to refine the fusion map by computing another residual fusion map $M{^{\prime}_{ab}}^{i}$. As illustrated in Fig.~\ref{fig:homo_decoder_vis}, in order to produce $M{^{\prime}_{ab}}^{i}$, feature maps $F^{i}_{a}$ and $F^{i}_{b}$ are firstly multiplied by the Fusion Map $M_{ab}^{i}$, $M_{ba}^{i}$ as:
\begin{equation}
    F^{{\prime}^{i}}_{a} = F^{i}_{a} * M_{ab}^{i},
    F^{{\prime}^{i}}_{b} = F^{i}_{b} * M_{ba}^{i},
\end{equation}
after computing the masked feature maps $F^{{\prime}^{i}}_{a}$ and $F^{{\prime}^{i}}_{b}$, they are fed into another Map Block which is the same as in SGF, yielding a refine fusion map $M{^{\prime}_{ab}}^{i}$ ranging from $\gamma_{-}$ to $\gamma_{+}$. Given the initial fusion map $M_{ab}^{i}$ and the refine fusion map $M{^{\prime}_{ab}}^{i}$, the final fusion map can be computed by $M_{ab}^{i} + M{^{\prime}_{ab}}^{i}$.

\textbf{Homo and Fuse Blocks.} To further address the problem of gyro field, we propose Homo and Fuse Blocks which produce a motion field to combine with the gyro field, yielding a complete homography. The design of the module follows the BasesHomo~\cite{ye2021motion}. More specifically, due to the rolling-shutter effect, as the gyro field is consist of an array of rotational-only homography, the output motion field is also composed of a mixture model which is smoothed along the rolling-shutter direction. In the Homography Decoder, based on the gyro field and feature maps, the output homography can be converted into a motion field to replace the initial gyro field for subsequent training. To summarize, we find that the learning of optical flow and deep homography can be mutually reinforcing for giving gyroscope data with our method.

\section{Experimental Results}
\subsection{Dataset}
\subsubsection{Optical Flow}{
The representative datasets for optical flow estimation and evaluation include FlyingChairs~\cite{dosovitskiy2015flownet}, MPI-Sintel~\cite{butler2012naturalistic}, KITTI 2012~\cite{geiger2012we}, KITTI 2015~\cite{menze2015object} and GOF~\cite{li2021gyroflow}.

    \textbf{FlyingChairs.} The ``Flying Chairs"~\cite{dosovitskiy2015flownet} is a synthetic dataset with optical flow ground truth. It consists of $22872$ image pairs with the size of $384$ x $512$ and corresponding flow fields for training, $640$ image pairs, and corresponding flow fields for testing. Images show renderings of 3D chair models moving in front of random backgrounds from Flickr. The motions of both the chairs and the background are purely planar.
    
    \textbf{MPI-Sintel.} The ``MPI-Sintel"~\cite{butler2012naturalistic} is a synthetic dataset for the training and evaluation of optical flow derived from the open source 3D animated short film, Sintel. It consists of $1041$ image pairs with the size of $436$ x $1024$ for training and $564$ image pairs for testing. Moreover, it provides three different rendering passes named ``Sintel-Albedo", ``Sintel-Clean", and ``Sintel-Final". The ``Sintel-Albedo" pass has constant brightness in most regions, the ``Sintel-Clean" pass introduces illumination, and the ``Sintel-Final" introduces motion blur, focus blur, and atmospheric effects.
    
    \textbf{KITTI.} The ``KITTI 2012"~\cite{geiger2012we} and the ``KITTI 2015"~\cite{menze2015object} are realistic datasets for the tasks of stereo, optical flow, visual odometry / SLAM, and 3D object detection which are developed by the autonomous driving platform equipped with four high-resolution video cameras, a Velodyne laser scanner and a state-of-the-art localization system. It contains 194 training pairs and 195 test pairs whose image size is $376$x$1240$. The ``KITTI 2015"~\cite{menze2015object} contains 200 training pairs and 200 test pairs.
    
    \textbf{GOF.} The ``GOF"~\cite{li2021gyroflow} is the first dataset combining gyroscope data with video frames in the case of cameras without an optical image stabilizer. It is collected in different seasons and different environments, such as foggy scenes in the winter and rainy scenes in the summer. There are $4$ different categories, including regular scenes (RE), low light scenes (Dark), foggy scenes (Fog), and rainy scenes (Rain). It contains about $5,000$ pairs for the training and $280$ pairs for the evaluation.}

\subsubsection{Homography}{
The representative datasets for homography estimation include MegaDepth~\cite{li2018megadepth}, HPatches~\cite{balntas2017hpatches}, CAHomo~\cite{zhang2020content} and GF4~\cite{li2021deepois}. All the datasets support the training and evaluation of deep homography estimation.

    \textbf{MegaDepth.} MegaDepth~\cite{li2018megadepth} is generated by leveraging multi-view internet photo collections via modern structure-from-motion (SFM) and multi-view stereo (MVS) methods. The images show extreme viewpoint and appearance variations with sparse ground truth. 
    
    \textbf{HPatches.} HPatches~\cite{balntas2017hpatches} are extracted from a number of image sequences, where each sequence contains images of the same scenes with illumination and viewpoint changes from minor to major.
    
    \textbf{CAHomo.} CAHomo~\cite{zhang2020content} is the first to propose to evaluate the homography estimation in challenging scenarios that contains 5 categories, including regular (RE), low-texture (LT), low-light (LL), small-foregrounds (SF), and large-foregrounds (LF) image pairs that each category contains around 16k image pairs.
    
    \textbf{GF4.} GF4~\cite{li2021deepois} is the first dataset that combines gyroscope data and video frames in the case of optical image stabilizer cameras. Similar to CAHomo~\cite{zhang2020content}, it consists of $4$ categories, regular(RE), low-texture(LT), low-light(LL) and moving-foreground(MF). Each category contains $350$ pairs, a total of $1400$ pairs, along with synchronized gyroscope readings.
}

\begin{figure*}[t]
\begin{center}
  \includegraphics[width=1\linewidth]{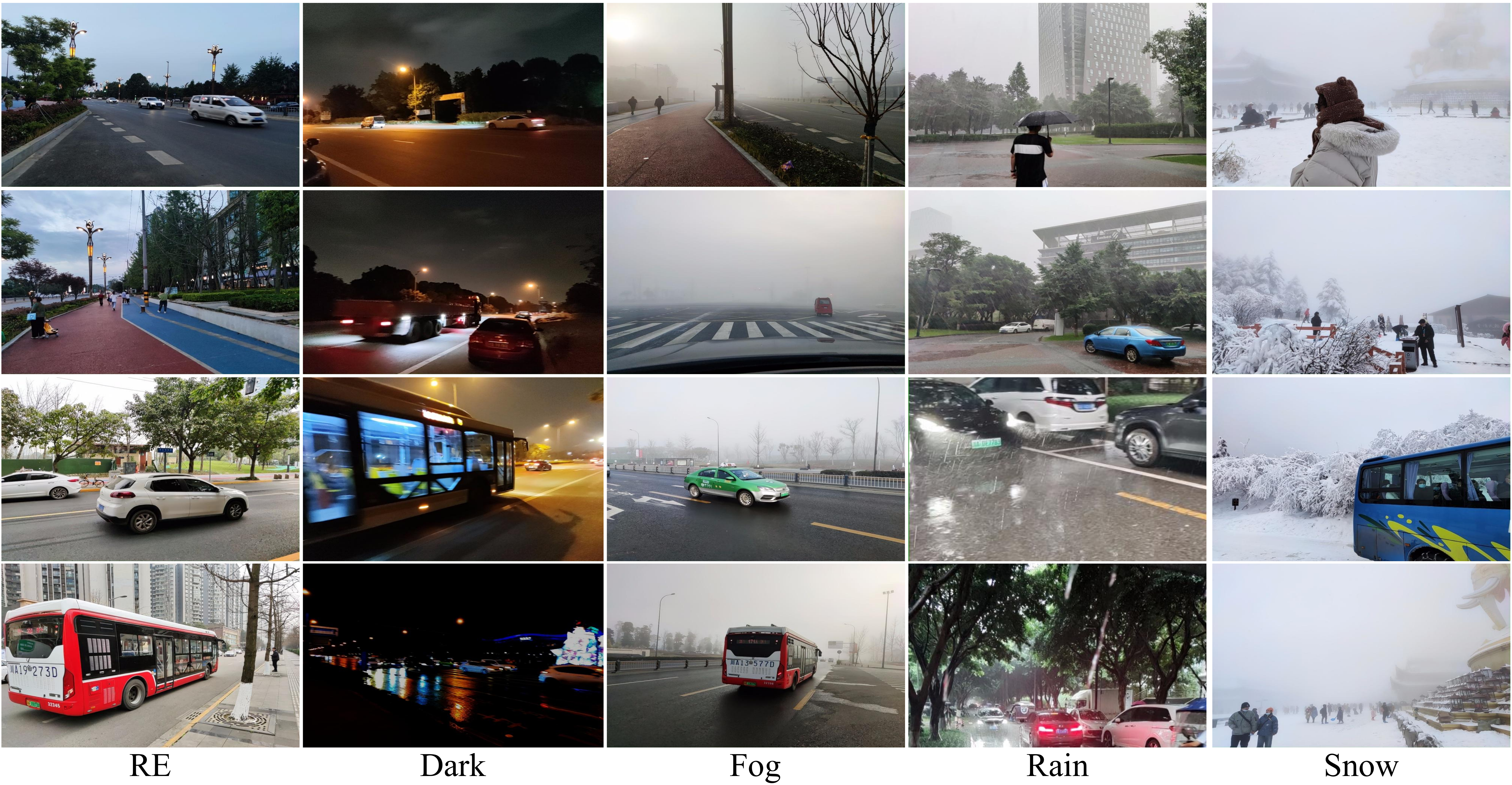}
\end{center}
  \caption{A glance at our dataset which can be divided into $5$ categories, including regular scenes(RE), low light scenes(Dark), foggy scenes(Fog), rainy scenes(Rain), and snowy scenes(Snow). The dataset is collected in different environments and different seasons, such as foggy scenes in the winter, rainy scenes in the summer, and snowy scenes in the mountains. Each category contains about $2,000$ pairs for training and $140$ pairs for evaluation, so a total of more than $10,000$ pairs dataset is proposed with synchronized gyroscope readings.}
\label{fig:dataset}
\end{figure*}

\begin{figure}[h]
\begin{center}
  \includegraphics[width=1\linewidth]{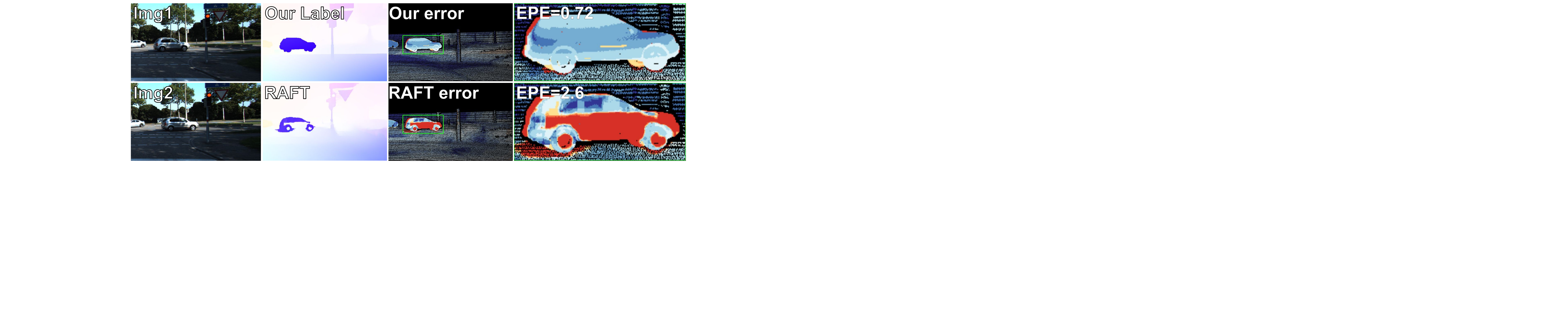}
\end{center}
  \caption{One label example on KITTI 2012~\cite{geiger2012we}, compared to RAFT\cite{teed2020raft}(the second line) that computes an EPE equals $2.6$, our label flow(the first line) produces a $0.72$ EPE. From the error map, we notice that our labeled optical flow is much more accurate.}
\label{fig:liuce_gt}
\end{figure}

\begin{figure}[t]
\begin{center}
  \includegraphics[width=1\linewidth]{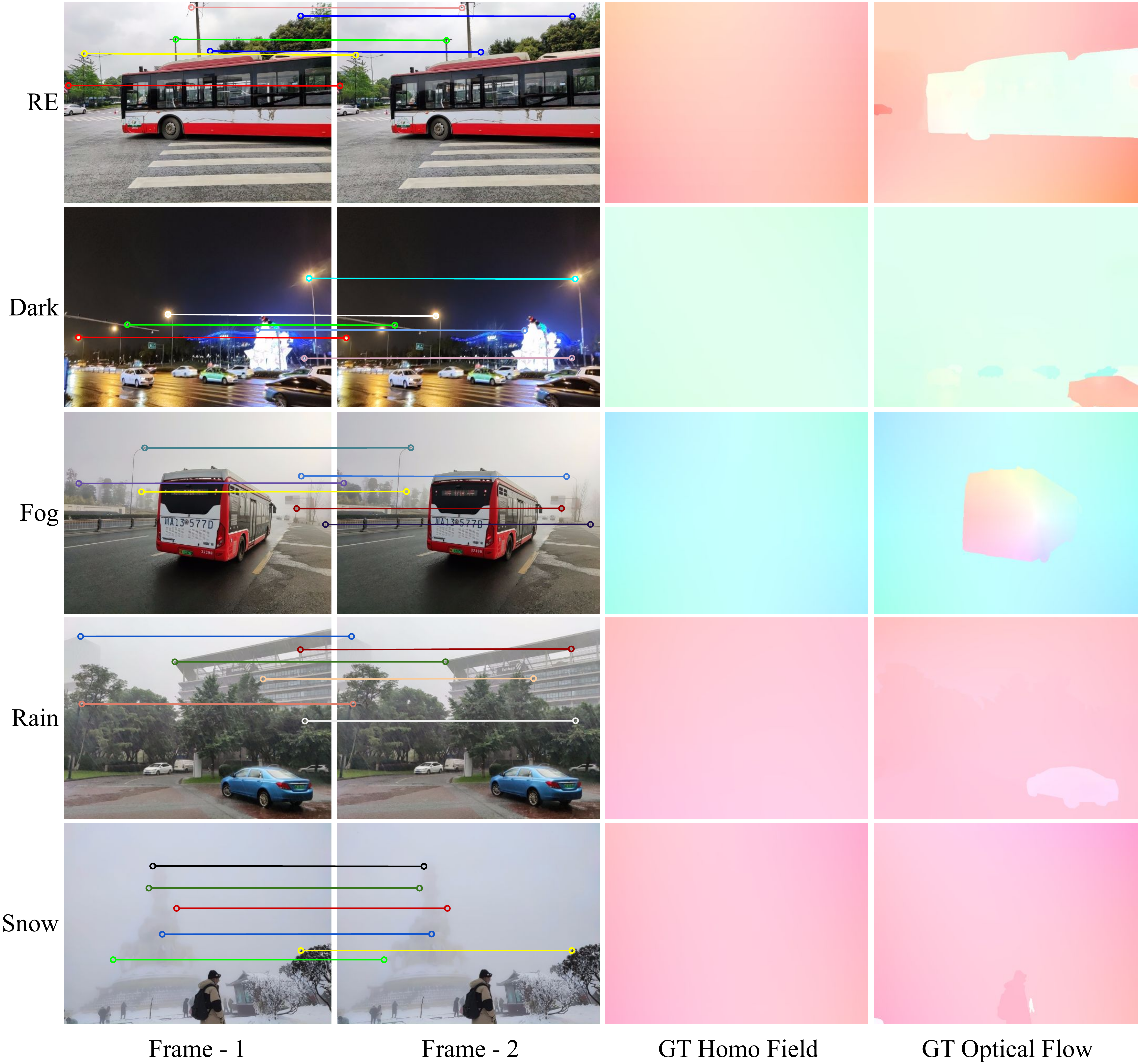}
\end{center}
  \caption{A glance at the GHOF benchmark which can be divided into $5$ categories, including regular scenes(RE), low light scenes(Dark), foggy scenes(Fog), rainy scenes(Rain) and snowy scenes(Snow). Each category contains about $100$ pairs. The benchmark is capable of evaluating homography and optical flow estimation. The first and second columns are consecutive frames, a batch of uniformly matching points are manually selected as GT correspondences, then producing the ground-truth homography field as shown in the third column. We leverage the method of ~\cite{liu2008human} to annotate the ground-truth optical flow as illustrated in the last column. A total of $530$ pairs evaluation dataset is proposed with gyroscope readings.}
\label{fig:benchmark}
\end{figure}

\subsubsection{GHOF Dataset}{
Currently, GOF~\cite{li2021gyroflow} is the only dataset that supports the training and evaluation for gyroscope-based optical flow estimation. On the other hand, although GF4~\cite{li2021deepois} provides gyroscope data and sparse correspondence ground truth, it can not be used for the gyroscope-based homography estimation tasks due to the existence of the OIS(optical image stabilizer) module. which breaks the correlation between image and gyroscope~\cite{li2021deepois}. In order to facilitate the learning and evaluation of the optical flow and homography estimation, we implement the GOF~\cite{li2021gyroflow} dataset and propose a new dataset named \textbf{GHOF}.

\textbf{Training Set.} A set of videos with gyroscope readings are recorded using a cellphone. Compared to GF4~\cite{li2021deepois}, which uses a phone with an OIS camera. We carefully choose a non-OIS camera phone to eliminate the effect of the OIS module. We collect videos in $5$ different environments and seasons as illustrated in Fig.~\ref{fig:dataset}, including regular scenes (RE), low light scenes (Dark), foggy scenes (Fog) in the winter, rainy scenes (Rain) in the summer, and snowy scenes (Snow) in the mountains. Notably, the most challenging cases during collecting are fog, rain, and snow because they contain both veiling and streaks effect~\cite{li2019rainflow}, where the streaks effect is non-trivial to be photographed. As a result, looking and waiting for the appropriate opportunities(heavy fog, rain, and snow) cost us a lot of time and effort. Moreover, to address the rotational camera motion and the foreground rigid motion problems, we film data with parallax variations and human movement, then the new data is added or replaced to the original dataset. Specifically, for each scenario, we record multiple videos lasting from $30$ to $60$ seconds, yielding around $2,000$ frames under every environment. In total, we collect around $10,000$ frames for the training set.

\textbf{Evaluation Set.}{
For quantitative evaluation, the ground-truth optical flow and sparse correspondences are required for each pair. We firstly collect $254$ samples under less challenging environments, such as slight mist, light rain, and nightfall. Furthermore, we capture the most challenging scenes or add some effects manually to the input frames which do not change the corresponding matching relation, such as noise, rain streaks, snow streaks, and focus blur which includes $276$ pairs, finally yielding
a $530$ pairs evaluation set. Fig.~\ref{fig:benchmark} shows examples.

    \textbf{Optical Flow.} For the optical flow part, labeling ground-truth flow is non-trivial. As far as we know, no powerful tool is available for this task. Following~\cite{li2019rainflow, yan2020optical}, we adopt the most related approach~\cite{liu2008human} to label the ground-truth flow with the best efforts. It costs approximately $20\sim30$ minutes per image, especially for challenging scenes. We firstly label an amount of $600$ examples containing rigid and non-rigid objects, then we select those with good visual performance, i.e., the performance of image alignment, and discard the others. Furthermore, we refine the selected samples with detailed modifications around the motion boundaries. To verify the effectiveness of our labeled optical flow, we choose to label several samples from KITTI 2012~\cite{geiger2012we}. Given the ground-truth, we compare our labeled optical flow with results produced by the previous state-of-the-art supervised method, i.e., RAFT~\cite{teed2020raft}. Our labeled flow computes an endpoint error (EPE) of $0.7$, where RAFT computes an EPE of $2.4$, which is more than $3$ times larger than ours. Fig.~\ref{fig:liuce_gt} shows one example. As the error map illustrates, our labeled flow is much more accurate and complete than the SOTA method. We leverage this approach to generate optical flow ground-truth for evaluation. 
    
    \textbf{Homography.} For the homography part, we follow the pipeline developed in CAHomo~\cite{zhang2020content} and GF4~\cite{li2021deepois}, we firstly detect key points and match the correspondences between the image pair, then we perform the outlier reject algorithm such as RANSAC to filter the outliers. Next, given the correspondences, we manually select $5$ - $8$ pairs to compute a homography that can be used to warp the reference frame. Based on alignment results between the reference frame and the warping result, the ground truth sparse correspondences can be generated. We illustrate some examples in Fig.~\ref{fig:benchmark}, the visualization of correspondences are colored lines in the first and second columns. The motion field generated by homography is shown in the third column
}

\subsection{Implementation Details}
We conduct experiments on the GHOF dataset. Our method is built upon the PWC-Net~\cite{sun2018pwc}. For the first stage, we train our model for $100$k steps without the occlusion mask. For the second stage, we enable the bidirectional occlusion mask~\cite{meister2018unflow}, the census loss~\cite{meister2018unflow}, and the spatial transform~\cite{liu2020learning} to fine-tune the model for about $300$k steps.

We collect videos with gyroscope readings using Qualcomm QRD equipped with Snapdragon 7150, which records videos in $600 \times 800$ resolution. We add random crop, random horizontal flip, and random weather modification (add fog and rain~\cite{imgaug}) during the training. We report the average endpoint error (EPE) in the evaluation set. The implementation is in PyTorch, and one NVIDIA RTX 2080 Ti is used to train our network. We use Adam optimizer~\cite{kingma2014adam} with parameters setting as $LR=1.0 \times 10^{-4}$, $\beta_{1}=0.9$, $\beta_{2}=0.999$, $\varepsilon=1.0 \times 10^{-7}$. The batch size is $4$. It takes $3$ days to finish the entire training process. On single $1080$ti, the time to generate an optical flow is $58$ms per frame. Same to previous work~\cite{jonschkowski2020matters, luo2021upflow}, we use the photometric loss and smooth term to train the network.}

\begin{figure*}[t]
\begin{center}
  \includegraphics[width=1\linewidth]{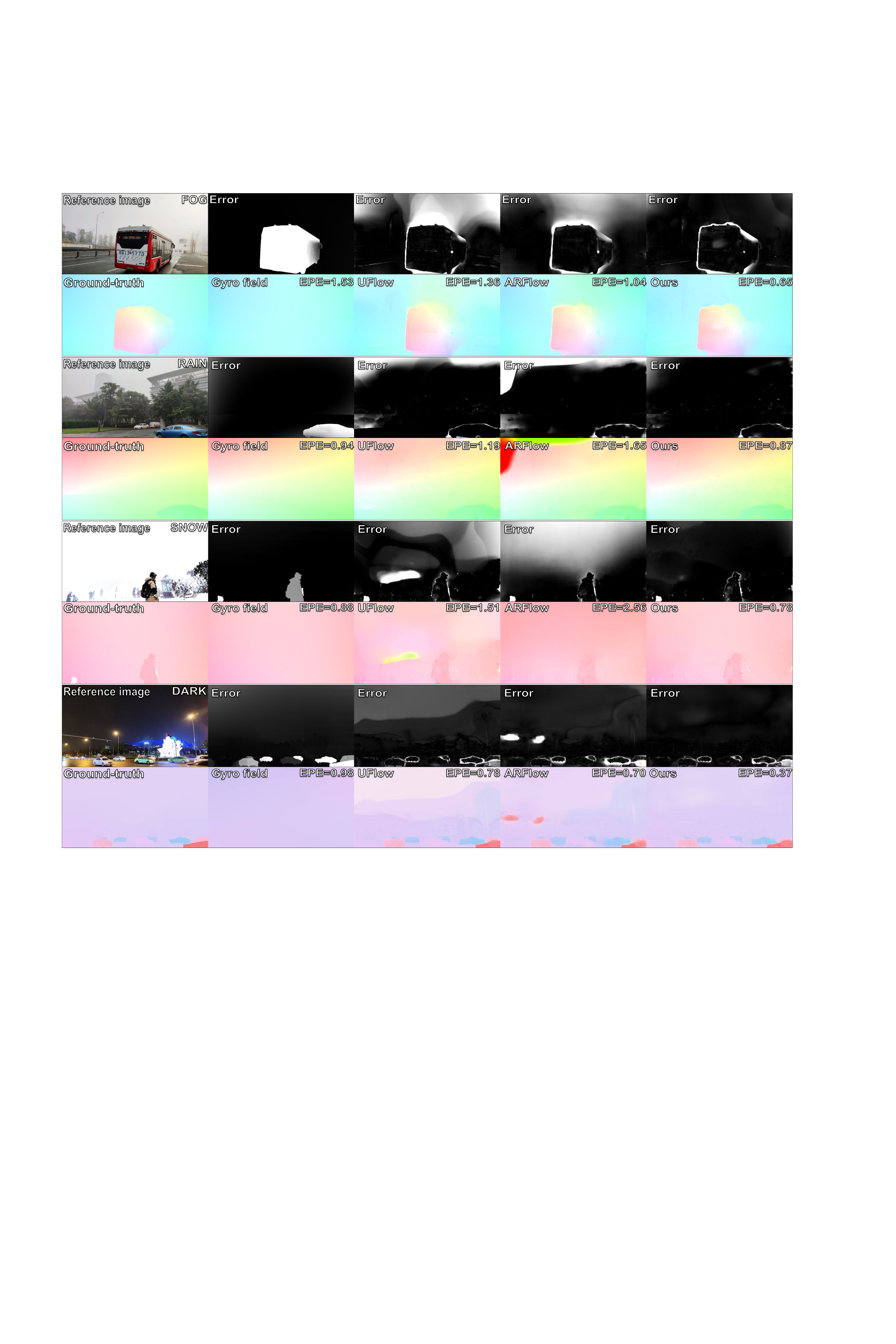}
\end{center}
  \caption{Visual comparison of our method with gyro field, ARFlow~\cite{liu2020learning}, and UFlow~\cite{jonschkowski2020matters} on the GHOF evaluation set. For the first $3$ challenging cases, we notice that our method achieves convincing results by fusing the background motion from the gyro field and the motion details from the optical flow. For the last example, in regular scenarios, fusing gyro field helps the learning of optical flow where the network produces accurate and sharp flow around the boundary of objects.}
\label{fig:qualitative}
\end{figure*}

\subsection{Comparisons with Image-based Methods}
In this section, we compare our method with traditional, supervised, self-supervised and unsupervised methods on the GHOF evaluation set with quantitative (Sec.~\ref{sec:quantitative}) and qualitative comparisons (Sec.~\ref{sec:qualitative}). To validate the effectiveness of key components, we conduct an ablation study in Sec.\ref{sec:ablation}. 

\subsubsection{Quantitative Comparisons\label{sec:quantitative}}

In Table~\ref{table:quantity}, the best results are marked in bold, and the second-best results are in underlining. `$I_{3\times3}$' refers to no alignment, and `Gyro Field' refers to alignment with pure gyro data. The Average EndPoint Error (AEPE), point matching errors (PME), and Percentage of Correct Keypoints (PCK) are chosen as evaluation metrics. AEPE is defined as the Euclidean distance between estimated and ground-truth optical flow, averaged over pixels of the target image. PCK-$1$ and PCK-$5$ are computed as the percentage of correspondences with a Euclidean distance error which is smaller than $1$ and $5$ respectively.

For traditional methods, we compare our GyroFlow+ with DIS~\cite{kroeger2016fast}, Farneback~\cite{farneback2003two} and DeepFlow~\cite{weinzaepfel2013deepflow} (Table~\ref{table:quantity}, (a)Tradition). As seen, their AEPEs are $3$ times larger than ours. In particular, DIS fails in snowy scenarios, and DeepFlow crashes in foggy and dark scenes. Moreover, we try to implement the traditional gyroscope-based optical flow method~\cite{li2018efficient} as no replies are received from the authors. Due to the lack of implementation details, we do not get reasonable results, so they are not reported.

Next, we compare with deep supervised optical flow methods, including Pwc-net~\cite{sun2018pwc}, RAFT~\cite{teed2020raft}, and recent state-of-the-art method GMA~\cite{jiang2021learning} (Table~\ref{table:quantity}, (b)Sup). For the lack of ground-truth labels during training, we cannot refine these methods on our trainset. So for each method, we choose models pre-trained on KITTI~\cite{geiger2012we} because the data distribution is closer to ours, containing both street view and moving cars. We later test them on the evaluation set. GMA performs the best, especially in rainy scenes, but it is not capable of handling the outliers, resulting in a low AEPE in the low-light category.

We refine self-supervised methods on our trainset, they do not rely on photometric loss and are optimized for homogeneous regions(Table~\ref{table:quantity}, (c)Self). However, their performances are still not as good as ours due to the effect of large-foreground-motion prevalent in the evaluation set.

We also compare our method to deep unsupervised optical flow methods, including UnFlow~\cite{meister2018unflow}, ARFlow~\cite{liu2020learning}, UFlow~\cite{jonschkowski2020matters} and UPFlow~\cite{luo2021upflow} (Table~\ref{table:quantity}, (d)Unsup).
Here, we refine the models on our training set. UPFlow achieves $4$ second-best results. While it is still not comparable with ours due to the challenging scenes.

We find that our GyroFlow+ model is robust in all scenes and computes a $1.19$ AEPE error and $95.82\%$ PCK-$5$px which are $26.54\%$ and $2.71\%$ better than the second-best method. Notably, for `Dark' scenes that consist of poor image texture, the `Gyro Field' alone achieves the second-best performance, indicating the importance of incorporating gyro motion, especially when the image contents are not reliable.

\begin{table*}[t]
    \small
    \centering
    \resizebox{\linewidth}{!}{
        \begin{tabular}{clcccccccccccc}
            \toprule
            & \multicolumn{1}{c}{} & \multicolumn{2}{c}{(\textbf{AVG})} & \multicolumn{2}{c}{(\textbf{RE})} & \multicolumn{2}{c}{(\textbf{FOG})} & \multicolumn{2}{c}{(\textbf{DARK})} & \multicolumn{2}{c}{(\textbf{RAIN})} & \multicolumn{2}{c}{(\textbf{SNOW})} 
            \\ 
            \cmidrule(lr){3-4} \cmidrule(lr){5-6} \cmidrule(lr){7-8} \cmidrule(lr){9-10} \cmidrule(lr){11-12} \cmidrule(lr){13-14} 
            &\multicolumn{1}{c}{\multirow{-2}{*}{Method}}&\emph{AEPE} &\emph{PCK-5px} &\emph{AEPE} &\emph{PCK-5px} &\emph{AEPE} &\emph{PCK-5px} &\emph{AEPE} &\emph{PCK-5px} &\emph{AEPE} &\emph{PCK-5px} &\emph{AEPE} &\emph{PCK-5px}
            \\ 
            
            \midrule
            {} & I33 & 6.47 & 41.10\% & 5.24 & 57.01\% & 7.67 & 18.41\% & 8.13 & 41.16\% & 5.26 & 50.71\% & 6.07 & 38.24\% \\
            
            {} & Gyro Field~\cite{li2021gyroflow} & 1.98 & 91.09\% & 3.29 & 75.78\% & 1.27 & 94.46\% & \underline{3.20} & \textbf{89.59\%} & 0.70 & \textbf{99.97\%} & 1.47 & 95.68\% \\
            
            {} & DIS~\cite{kroeger2016fast} & 5.68 & 75.94\% & 2.15 & 88.02\% & 5.18 & 66.86\% & 6.07 & 71.41\% & 1.44 & 91.12\% & 26.15 & 40.35\% \\
            
            {} & Farneback~\cite{farneback2003two} & 4.15 & 67.59\% & 2.4 & 84.76\% & 5.18 & 54.36\% & 7.11 & 55.67\% & 1.80 & 86.21\% & 4.79 & 55.28\% \\
            
            \multirow{-5}{*}{\rotatebox{90}{(a) Tradition}} & DeepFlow~\cite{weinzaepfel2013deepflow} & 6.20 & 48.12\% & 5.08 & 59.40\% & 7.38 & 22.11\%  & 8.14 & 48.17\% & 4.44 & 65.40\% & 6.48 & 38.41\% \\
            
            \midrule
            {} & Pwc-net~\cite{sun2018pwc} & 6.10 & 84.20\% & 2.60 & 90.71\% & 4.65 & 86.39\% & 6.34 & 77.08\% & 1.67 & 99.01\% & 15.25 & 67.81\% \\
            
            {} & RAFT~\cite{teed2020raft} & 6.01 & 83.95\% & 2.77 & 91.81\% & 5.87 & 86.71\% & 13.26 & 64.88\% & 0.88 & 98.08\% & 7.28 & 78.29\% \\
            
            \multirow{-3}{*}{\rotatebox{90}{(b) Sup}} & GMA~\cite{jiang2021learning} & 1.97 & 92.12\% & 1.31 & 94.22\% & 1.73 & 89.37\% & 4.87 & 79.28\% & 0.69 & 99.80\% & \underline{1.29} & \underline{97.26\%} \\
            
            \midrule
            {} & GLUNet~\cite{truong2020glu} & 6.86 & 62.81\% & 4.38 & 72.88\% & 6.16 & 58.05\% & 7.78 & 55.30\% & 3.73 & 80.08\% & 12.24 & 47.73\% \\
            
            {} & PDCNet~\cite{truong2021learning} & 7.07 & 77.24\% & 2.58 & 91.85\% & 5.02 & 74.39\% & 6.86 & 68.64\% & 2.34 & 91.85\% & 18.56 & 59.49\% \\
            
            \multirow{-3}{*}{\rotatebox{90}{{(c) Self}}}& WarpC~\cite{truong2021warp} & 5.96 & 69.49\% & 3.31 & 86.42\% & 5.96 & 61.49\% & 7.79 & 58.51\% & 2.84 & 87.77\% & 9.90 & 53.27\% \\
            
            \midrule
            {} & UnFlow~\cite{meister2018unflow} & 3.93 & 75.55\% & 2.51 & 83.95\% & 4.32 & 71.98\% & 6.77 & 62.21\% & 1.53 & 92.97\% & 4.52 & 66.65\% \\
            
            {} & ARFlow~\cite{liu2020learning} & 3.23 & 83.74\% & 2.07 & 89.38\% & 2.83 & 84.98\% & 5.36 & 73.20\% & 0.99 & 98.38\% & 4.9 & 72.74\% \\
            
            {} & UFlow~\cite{jonschkowski2020matters} & 1.66 & \underline{93.11\%} & 1.08 & 94.12\% & 1.28 & \underline{96.61\%} & 3.31 & 83.07\% & 0.66 & 99.25\% & 1.95 & 91.50\% \\
            
            \multirow{-4}{*}{\rotatebox{90}{(d) Unsup}} & UPFlow~\cite{luo2021upflow} & \underline{1.62} & 92.51\% & \underline{1.06} & \textbf{95.40\%} & \underline{1.20} & 95.87\% & 3.45 & 80.67\% & \underline{0.54} & 99.48\% & 1.88 & 91.12\% \\
            
            \midrule
             {} & Ours & \textbf{1.19} & \textbf{95.82\%} & \textbf{1.04} & \underline{94.57\%} & \textbf{1.14} & \textbf{96.86\%} & \textbf{2.20} & \underline{89.53\%} & \textbf{0.48} & \underline{99.86\%} & \textbf{1.12} & \textbf{98.27\%} \\
             
            \bottomrule
        \end{tabular}
    }
    \vspace{1mm}
    \caption{The Average EndPoint Error (AEPE) and Percentage of Correct Keypoints (PCK) of our method and all comparison methods on the GHOF benchmark. The best results are highlighted in bold, and the second best results are underlined. PCK-$5$px represents the percentage of correspondences with a Euclidean distance error that is smaller than $5$.}
    \label{table:quantity}
\end{table*}

\begin{table*}[h]
    \small
    \centering
    \resizebox{\linewidth}{!}{
        \begin{tabular}{clcccccccccccc}
            \toprule
            & \multicolumn{1}{c}{} & \multicolumn{2}{c}{(\textbf{AVG})} & \multicolumn{2}{c}{(\textbf{RE})} & \multicolumn{2}{c}{(\textbf{FOG})} & \multicolumn{2}{c}{(\textbf{DARK})} & \multicolumn{2}{c}{(\textbf{RAIN})} & \multicolumn{2}{c}{(\textbf{SNOW})} 
            \\ 
            \cmidrule(lr){3-4} \cmidrule(lr){5-6} \cmidrule(lr){7-8} \cmidrule(lr){9-10} \cmidrule(lr){11-12} \cmidrule(lr){13-14} 
            &\multicolumn{1}{c}{\multirow{-2}{*}{Method}}&\emph{PME} &\emph{PCK-1px} &\emph{PME} &\emph{PCK-1px} &\emph{PME} &\emph{PCK-1px} &\emph{PME} &\emph{PCK-1px} &\emph{PME} &\emph{PCK-1px} &\emph{PME} &\emph{PCK-1px}
            \\ 
            \midrule
            (1)& I33 & 6.33 & 2.44\% & 4.94 & 3.61\% & 7.24 & 0.00\% & 8.09 & 2.45\% & 5.48 & 0.85\% & 5.89 & 5.29\% \\
            (2)& Gyro Field~\cite{li2021gyroflow} & \underline{0.54} & \underline{84.59\%} & \underline{0.33} & \underline{97.73\%} & \underline{0.27} & \textbf{99.14\%} & \underline{1.06} & {55.55\%} & \textbf{0.44} & 88.64\% & \underline{0.57} & \underline{81.87\%} \\
            \midrule
            (3)& SIFT~\cite{lowe2004distinctive} & 4.31 & 51.09\% & 0.66 & 79.65\% & 3.95 & 60.08\% & 11.38 & 32.85\% & 0.64 & 79.92\% & 4.94 & 14.94\% \\
            (4)& ORB~\cite{rublee2011orb} & 15.14 & 16.24\% & 6.92 & 12.04\% & 31.27 & 1.88\% & 27.80 & 2.67\% & 1.82 & 49.70\% & 7.90 & 12.10\% \\
            (5)& BEBLID~\cite{suarez2020beblid} & 13.62 & 18.44\% & 5.91 & 15.87\% & 20.52 & 4.37\% & 32.02 & 3.98\% & 1.78 & 52.06\% & 7.89 & 13.49\% \\
            \midrule
            (6)& SOSNet~\cite{tian2019sosnet} & 5.29 & 47.39\% & 0.72 & 76.18\% & 5.56 & 38.81\% & 12.95 & 26.80\% & 0.77 & 74.75\% & 6.37 & 26.09\% \\
            (7)& SuperPoint~\cite{detone2018superpoint} & 4.22 & 42.45\% & 3.99 & 19.24\% & 1.94 & 60.91\% & 10.11 & 23.47\% & 0.73 & 75.56\% & 4.36 & 35.22\% \\
            (8)& Loftr~\cite{sun2021loftr} & 3.04 & 45.96\% & 2.64 & 29.31\% & 1.33 & 60.91\% & 6.71 & 30.68\% & 0.56 & 86.73\% & 3.94 & 26.60\% \\
            \midrule
            (9)& Supervised~\cite{detone2016deep} & 6.61 & 2.59\% & 6.04 & 3.38\% & 6.02 & 1.63\% & 7.68 & 2.14\% & 6.99 & 1.09\% & 6.32 & 4.33\% \\
            (10)& Unsupervised~\cite{nguyen2018unsupervised} & 6.33 & 2.29\% & 4.88 & 3.59\% & 7.42 & 1.12\%  & 8.11 & 2.07\% & 5.28 & 1.63\% & 5.96 & 3.60\% \\
            (11)& CAHomo~\cite{zhang2020content} & 3.87 & 20.33\% & 4.10 & 8.48\% & 3.84 & 11.31\%  & 6.99 & 21.54\% & 1.27 & 46.95\% & 3.17 & 11.53\% \\
            (12)& BasesHomo~\cite{ye2021motion} & 2.28 & 45.79\% & 2.02 & 36.02\% & 1.43 & 58.78\%  & 4.90 & 37.48\% & 0.78 & 73.80\% & 2.29 & 27.58\% \\
            (13)& HomoGAN~\cite{hong2022unsupervised} & 1.95 & 71.23\% & 1.73 & 64.20\% & 0.60 & 88.92\%  & 3.95 & \textbf{59.20\%} & 0.47 & \underline{95.21\%} & 3.02 & 54.40\% \\
            \midrule
             & Ours & \textbf{0.51} & \textbf{87.52\%} & \textbf{0.29} & \textbf{99.12\%} & \textbf{0.23} & \underline{98.96\%} & \textbf{1.04} & \underline{59.12\%} & \underline{0.46} & \textbf{95.33\%} & \textbf{0.56} & \textbf{85.06\%}
            \\ 
            \bottomrule
        \end{tabular}%
    }
    \vspace{1mm}
    \caption{The point matching errors (PME) and Percentage of Correct Keypoints (PCK) of our method and all comparison methods on the GHOF benchmark. The best results are highlighted in bold, and the second best results are underlined. PCK-$1$px represents the percentage of correspondences with a Euclidean distance error that is smaller than $1$.}
    \label{table:homo}
\end{table*}

In Table~\ref{table:homo}, we evaluate our homography with other methods which can be split into traditional feature-based methods(Table~\ref{table:homo}, $3$~-$5$),  learned feature-based methods(Table~\ref{table:homo}, $6$~-$8$), and deep homography estimation methods(Table~\ref{table:homo}, $9$~-$13$) on the first $256$ pairs in GHOF benchmark because the last $276$ samples are too difficult for comparison methods. The first $2$ categories use RANSAC~\cite{fischler1981random} as the outlier rejection algorithm. Our method achieves state-of-the-art performance in $10$ terms and produces the second best results in the rest $2$ terms. In the regular scenes(RE), SIFT~\cite{lowe2004distinctive} achieves the third best result because while there are sufficient high-quality feature points, the dynamic foregrounds especially in large sizes disturb the matching results. Our result is $12$\% better than the gyro field as the translation component is complemented. As for challenging cases, our method is at least $20.69\%$ and at most $78.78\%$ better than the feature-based or deep homography methods. Moreover, the gyro field achieves the second best in most scenarios, indicating the robustness and importance of gyroscope data. However, our results outperform the gyro field with the help of translation information. 

\subsubsection{Qualitative Comparisons\label{sec:qualitative}}

In Fig.~\ref{fig:qualitative}, we illustrate the qualitative results of the evaluation set. We choose one example for each of four different challenging scenes, including the low-light scene (Dark), the foggy scene (Fog), the rainy scene (Rain), and the snowy scene (Snow). For comparing methods, we choose the gyro field and $2$ recent unsupervised methods, i.e., ARFlow~\cite{liu2020learning} and UFlow~\cite{jonschkowski2020matters} which are refined on our training set. In Fig.~\ref{fig:qualitative}, we show optical flow along with corresponding error maps and also report the AEPE error for each example. As shown, for challenge cases, our method can fuse the background motion from the gyro field with the motion of dynamic objects from the image-based optical flow, delivering both better visual quality and lower AEPE errors. While the $2$ image-based methods fail in the homogeneous region, typical examples of which are shown, including foggy scenes where the sky is occluded by fog(Fig.~\ref{fig:qualitative}, row $1$~-$2$), rainy scenes where buildings are obscured by rain-veiling(Fig.~\ref{fig:qualitative}, row $3$~-$4$), snowy scenes where the background is hidden by snow-veiling(Fig.~\ref{fig:qualitative}, row $5$~-$6$), and low-light scenes where the backdrop is covered by night(Fig.~\ref{fig:qualitative}, row $7$~-$8$).

The unsupervised optical flow methods~\cite{liu2019ddflow,liu2020learning,jonschkowski2020matters} are supposed to work well in RE scenes given sufficient texture. However, we notice that, even for the RE category, our method outperforms the others, especially at the motion boundaries as shown in Fig.~\ref{fig:SGF-quatitative}. With the help of the gyro field that solves the global motion, the network can focus on challenging regions. As a result, our method still achieves better visual quality and produces lower AEPE errors in RE scenarios.

\subsection{Ablation Studies\label{sec:ablation}}

  \begin{table}[!h]
    \centering
    \resizebox{0.99\linewidth}{!}{  
      \begin{tabular}{rlllllll}
        \toprule
        1) & Modification &AVG & RE & FOG & DARK & RAIN & SNOW \\
        \midrule
        2) & $1/32$ resolution & 1.48 & 1.13 & 1.16 & 3.09 & 0.53 & 1.52 \\ 
        3) & $1/16$ resolution & 1.36 & 1.06 & 1.10 & 2.67 & 0.54 & 1.44\\ 
        4) & $1/8$ resolution & 1.52 & 1.19 & 1.25 & 3.15 & 0.61 & 1.41\\ 
        5) & $1/4$ resolution & 1.60 & 1.05 & 1.28 & 3.20 & 0.58 & 1.91\\
        \midrule
        6) & Middle & 1.30 & 1.21 & 1.05 & 2.62 & 0.50 & \underline{1.10} \\ 
        7) & Tail & 1.37 & 1.10 & 1.07 & 2.84 & 0.54 & 1.32 \\ 
        \midrule
        8) & DWI & 1.63 &  1.16  & 1.26 & 3.21 & 0.63 & 1.92 \\ 
        9) & DPGF & 1.42 & 1.08 & 1.23 & 2.69 & 0.56 & 1.54 \\ 
        10) & SGF-Fuse &  1.46 & 1.05 & 1.13 & 3.00 & 0.58 & 1.53
        \\ 
        11) & SGF-Map &  1.50 & 1.25 & 1.37 & 2.79 & 0.51 & 1.60\\ 
        12) & SGF-Dense &  1.30 & 1.32 & \underline{0.99} & 2.67 & \underline{0.49} & \textbf{1.03}
        \\
        \midrule
        13) & UFlow + SGF & \underline{1.22} & \textbf{1.03} & \textbf{0.95} & \underline{2.44} & 0.55 & 1.13 \\
        \midrule
        14) & GyroFlow & 1.31 & 1.21 & 1.06 & 2.63 & \underline{0.49} & 1.17 \\
        \midrule
        15) & Ours & \textbf{1.19} & \underline{1.04} & 1.14 & \textbf{2.20} & \textbf{0.48} & 1.12\\
        \bottomrule
    \end{tabular}}
        \vspace{1mm}
    \caption{Results of ablation studies. The Average EndPoint Error (AEPE) is reported where each row is the result of our method with a specific modification, please refer to the text.} 
    \label{tab:ablation}
  \end{table}


To evaluate the effectiveness of the design for each module, we conduct ablation experiments on the evaluation set. AEPE errors are reported under $5$ categories, including Dark, Fog, Rain, and RE, along with the average error.

\subsubsection{Gyro Field Fusion Layer}
Intuitively, it is possible to fuse the gyro field only once during the training, so we add our SGF module at a specific pyramid layer. As illustrated in Table~\ref{tab:ablation} 2~-5, we notice that the more bottom layer we add SGF to, the lower EPE error it produces, and the $1/16$ resolution, i.e., the second pyramid layer achieves the best results. From our experiments, we find that each pyramid layer is responsible for a different motion, with the top layer more inclined to generate the global motion and the bottom layer more inclined to fine-tune local motion. So giving the gyro field at the top level helps the estimation of optical flow. However, the best results can only be obtained when we add the gyro field at all layers.

\subsubsection{Gyro Field Fusion Position}
Our decoder is based on Pwc-net~\cite{sun2018pwc}, so the optical flow can be generated in three positions. The first position is at the head of the decoder, where the optical flow from the previous layer is upsampled to the next layer. The second position is at the middle of the decoder, where optical flow is refined by the correlation layer, i.e., cost volume~\cite{sun2018pwc}. The last position is at the tail of the decoder, where optical flow is fine-tuned by the context network~\cite{sun2018pwc}. As illustrated in Table~\ref{tab:ablation} 6~-7, we notice that the later 
SGF is positioned, the worse the result is, and the head position achieves the best results.

\subsubsection{The Design of SGF}

For SGF, we test several designs and report results in Table~\ref{tab:ablation} 8~-12. First of all, two straightforward methods are adopted to build the module. DWI refers that we directly warp the $I_{a}$ with the gyro field, then we input the warped image and $I_{b}$ to produce a residual optical flow. DPGF denotes that, for each pyramid layer, we directly add the gyro field onto the optical flow. As shown in Table~\ref{tab:ablation}, for DWI, the result is not good. Except for the absence of gyroscope guidance during training, another possibility is that the warping operation breaks the image structure such as blurring and noising. DPGF gets a better result but is still not comparable to our SGF design because the gyro field registers background motion that should not be concatenated to dynamic object motion. Furthermore, we compare our SGF with three variants: (1) SGF-Fuse, we remove the map block and the final fusion procedure. Although it computes a $1.46$ AEPE error, it performs unstable in challenging scenes; (2) SGF-Map, where the fusion block is removed. It results in worse performance because the fusion map $M_{ab}$ tends to be inaccurate. (3) SGF-Dense, we integrate the two blocks into one unified dense block, which produces a $3$ channels tensor of which the first two channels represent the fusion flow $O_{ab}$, and the last channel denotes the fusion map $M_{ab}$. Our SGF is much better on average.

\subsubsection{Unsupervised Methods with SGF} 

\begin{figure}[h]
\begin{center}
  \includegraphics[width=0.98\linewidth]{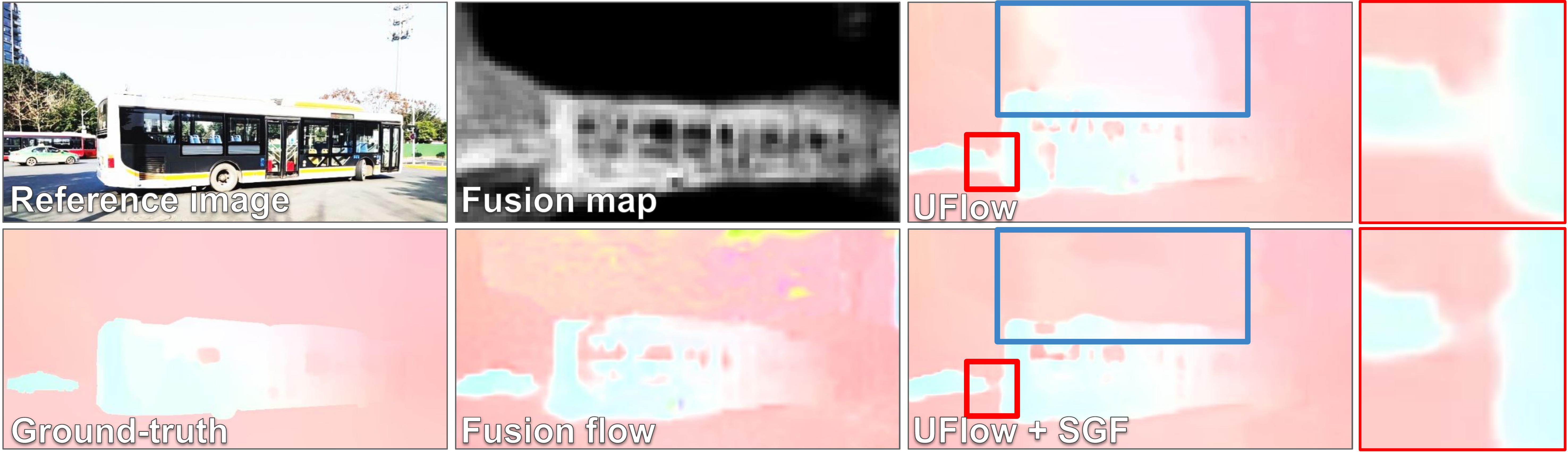}
\end{center}
 \caption{Visual example of our self-guided fusion module(SGF). Results of UFlow and UFlow with SGF are shown. The fusion map is used to guide the network to focus on motion details.}
\label{fig:SGF-quatitative}
\end{figure}

We insert the SGF module into an unsupervised method~\cite{jonschkowski2020matters}. In particular, similar to Fig.~\ref{fig:pipeline}, we add the SGF before the decoder $\mathbf{D}$ for each pyramid layer. The unsupervised method is trained on our dataset, and we report AEPE errors in Table~\ref{tab:ablation} row $13$. After inserting our SGF module, noticeable improvements can be observed in Table~\ref{table:quantity} and Table~\ref{tab:ablation}, which proves the effectiveness of our proposed SGF module. Fig.~\ref{fig:SGF-quatitative} shows an example. Both background motion and boundary motion are improved after integrating our SGF. Lastly, we compare GyroFlow+ with the original GyroFlow, with the help of the constrained explicit fusing strategy and the homography decoder, distinct improvements can be observed especially in the average and regular scenarios.

\section{Conclusion}
We have presented a novel framework GyroFlow+ for unsupervised homography and optical flow learning by fusing the gyroscope data which is the first to bring gyroscope, homography and optical flow together. We have proposed a self-guided fusion module to fuse the gyro field and optical flow, and a homography decoder to implement gyro field to homography. For the evaluation, we have proposed a dataset GHOF and labeled $500$+ ground-truth optical flow for quantitative metrics. The results show that our proposed method achieves state-of-the-art in all regular and challenging categories compared to the existing methods. Code and dataset will be available to facilitate future research.

{
\small
\bibliographystyle{ieeetr}
\bibliography{pami_bib}
}

\end{document}